# EVALUATION OF DISTANCE-BASED APPROACHES FOR FORENSIC COMPARISON: APPLICATION TO HAND ODOR EVIDENCE

**Authors**: Isabelle Rivals (PhD) [a], Cédric Sautier (MS) [b], Guillaume Cognon (MS) [b], Vincent Cuzuel (phD) [b]

[a] Equipe de Statistique Appliquée, ESPCI Paris, PSL Research University, INSERM, UMRS 1158 Neurophysiologie Respiratoire Expérimentale et Clinique, 10 rue Vauquelin, Paris, France

[b] Institut de Recherche Criminelle de la Gendarmerie Nationale, Caserne Lange, 5 boulevard de l'Hautil, BP 20312 Pontoise, 95037 Cergy Pontoise CEDEX, France

**Corresponding author**: Isabelle Rivals
Address: Equipe de Statistique Appliquée, ESPCI Paris, 10 rue Vauquelin, 75005 Paris, France
E-mail: isabelle.rivals@espci.fr

**Short version of title (running head)**: Towards odor-based forensic identification

# EVALUATION OF DISTANCE-BASED APPROACHES FOR FORENSIC COMPARISON: APPLICATION TO HAND ODOR EVIDENCE

**ABSTRACT**

The issue of distinguishing between the same-source and different-source hypotheses based on various types of traces is a generic problem in forensic science. This problem is often tackled with Bayesian approaches, which are able to provide a likelihood ratio that quantifies the relative strengths of evidence supporting each of the two competing hypotheses. Here, we focus on distance-based approaches, whose robustness and specifically whose capacity to deal with high-dimensional evidence are very different, and need to be evaluated and optimized.

A unified framework for direct methods based on estimating the likelihoods of the distance between traces under each of the two competing hypotheses, and indirect methods using logistic regression to discriminate between same-source and different-source distance distributions, is presented. Whilst direct methods are more flexible, indirect methods are more robust and quite natural in machine learning. Moreover, indirect methods also enable the use of a vectorial distance, thus preventing the severe information loss suffered by scalar distance approaches.

Direct and indirect methods are compared in terms of sensitivity, specificity and robustness, with and without dimensionality reduction, with and without feature selection, on the example of hand odor profiles, a novel and challenging type of evidence in the field of forensics. Empirical evaluations on a large panel of 534 subjects and their 1690 odor traces show the significant superiority of the indirect methods, especially without dimensionality reduction, be it with or without feature selection.

**HIGHLIGHTS**

• Direct and indirect distance-based likelihood ratio estimation methods for forensic comparison are investigated

• These methods are applied to high-dimensional evidence consisting of hand odor traces

• The methods' robustness, AUC, sensitivity and specificity are evaluated on a panel of 534 subjects

• Indirect methods based on logistic regression outperform direct ones and are more robust

• The indirect method using a vectorial distance outperforms that using a scalar one, both with and without feature selection

## 1. Introduction

A generic problem in courts of law is to decide whether a trace of an unknown origin, often drawn from a crime scene, and a specimen from a known source, stem from the same source, for example a person or a firearm. If the source is a person, the traces might be biometric such as a DNA profile [1-2], fingerprints [3], a voice [4], an olfactory profile [5], or they might consist of footwear impressions [6], handwriting [7], etc. If the source is a firearm, the traces may be features such as striations and impressions on a bullet or on a cartridge case [8].

The most common approach of forensic science to this problem is to estimate a likelihood ratio (LR), i.e. the ratio of the joint probability of occurrence of the two traces under the hypothesis that they arose from the same source and under the hypothesis that they arose from different sources. A convenient solution is to replace the joint probability of the traces by the probability of a distance between the two traces quantifying their dissimilarity [9, 10, 6, 11, 12, 13, 14]. If, as is most often the case, the distance is scalar, there is an important loss of information. Thus, we choose to focus on distance-based methods, but with the possibility to use a vectorial distance between traces.

Furthermore, the distance-based LR estimate can be obtained either directly, by estimating the distance likelihoods under the two hypotheses, or indirectly, by first using logistic regression to discriminate between same-source and different source distance distributions, and then Bayes' formula to infer the LR. Whilst the direct method is more flexible, the indirect method is more robust and quite natural in machine learning [15-16]. It is sometimes advocated for in the forensic context, for the same reasons and also because it enables score calibration and fusion with minimal mathematical complexity [17-18]. Here, we show that the indirect method also enables the use of a vectorial distance, thus preventing the severe information loss suffered by scalar distance approaches. We discuss the direct and indirect methods in terms of robustness and ability to handle high dimensional evidence, with or without dimensionality reduction, and with or without feature selection. We evaluate them in terms of sensitivity, specificity and robustness on the example of traces consisting of a hand odor profile.

## 2. Materials and methods

*2.1 Problem statement*

The aim is, given the evidence consisting of a pair of traces (e.g. two olfactory profiles), to decide whether these traces have the same source (e.g. the same person) or not. In the following, $H_{ss}$ refers to the hypothesis that the two traces stem from the same source, and $H_{ds}$ to the alternative hypothesis that they stem from different sources. Given the *a priori* probabilities $P(H_{ss})$ and $P(H_{ds})$, the Bayesian formula yields the posterior probability of $H_{ss}$ given the evidence E:

$$P(H_{ss}|E) = \frac{f(E|H_{ss})P(H_{ss})}{f(E|H_{ss})P(H_{ss}) + f(E|H_{ds})P(H_{ds})} \quad (1)$$

where $f(E|H_{ss})$ and $f(E|H_{ds})$ are the distributions of the evidence under $H_{ss}$ and $H_{ds}$, or likelihoods. Jeffreys developed an absolute scale to evaluate the degree of confidence in the same-source hypothesis outside a decisional framework based on the posterior probability of $H_{ss}$ using the LR [19-20] defined as:

$$LR(E) = \frac{f(E|H_{ss})}{f(E|H_{ds})} \quad (2)$$

which is independent of the *a priori* probabilities. In fact, the observed evidence E consists of the two traces which are represented by n-dimensional vectors (whose components are the amounts of each odor compound). Since we focus on distance-based methods, the information contained in these two vectors is transformed into a distance or dissimilarity measure, which can be either a scalar or an n-vector (a distance for each feature of the trace, here for each odor compound). In the following sections, this distance between the two traces will be denoted by *d*.

*2.2 Candidate methods*

The LR can be obtained either directly, i.e. by estimating the likelihoods of the distance under the two competing hypotheses, or indirectly, i.e. using first logistic regression to discriminate between same-source and different source distance distributions, and then formulas (1) and (2) to infer the LR.

2.2.1 Direct methods

For direct methods, we need to be able to evaluate $f(d|H_{ss})$ and $f(d|H_{ds})$ for any value of d. For this purpose, part of the available dataset can be used to build pairs of traces of the two types: same-source and different-source pairs. If d is scalar, or of dimension 2 or 3 at most, the two empirical distributions can be fitted, with Gaussian mixtures, for example, leading to parametric estimates of $f(d|H_{ss})$ and $f(d|H_{ds})$ [8, 10, 13]. In the case of many features, a fit of each component of d can be performed in the same way, and the overall likelihoods can be approximated through the product of the likelihoods in each dimension, leading to the naïve Bayes classifier. To be successful, the latter approach requires however that the features are not overly correlated.

2.2.2. Indirect methods

The aim of these method is to build a discriminative model of the boundary between the two categories of pairs (same-source and different-source pairs) rather than a generative model explicitly parameterizing the distributions in the two categories. It is well known that, under the hypothesis of single-Gaussian distributions with the same variance under $H_{ss}$ and $H_{ds}$ in the scalar case, or same covariance matrix in the multidimensional case, the posterior probability of $H_{ss}$ takes the form of a sigmoidal curve [15-16], hence the motivation for a logistic regression approach. Despite its result being a discriminative model, it enables to calculate the posterior probability of Equation (1) as well as the LR of Equation (2). As a matter of fact, the logistic regression model with parameters $\theta = [a^T\ b]^T$ has output:

$$r(d,\theta) = \frac{1}{1+\exp(-(a^T d + b))} \tag{3}$$

where *b* is a scalar, and *a* is either a scalar in the case of a scalar distance, or otherwise an n-vector (of the dimension of the evidence). If the proportions of the same-source and different-source categories in the calibration set are denoted by f$_{ss}$ and f$_{ds}$, r(d, θ) approximates:

$$\frac{f(d|H_{ss})f_{ss}}{f(d|H_{ss})f_{ss} + f(d|H_{ds})f_{ds}} \qquad (4)$$

Thus, the posterior probability for *a priori* probabilities P(H$_{ss}$) and P(H$_{ds}$) can be retrieved with:

$$P(H_{ss}|d) = \frac{1}{1 + \exp(-(a^T d + b))\frac{f_{ss}}{f_{ds}}\frac{P(H_{ds})}{P(H_{ss})}} \qquad (5)$$

and the likelihood ratio with:

$$LR(d) = \exp(a^T d + b)\frac{f_{ds}}{f_{ss}} \qquad (6)$$

*2.3. Pros and cons*

The indirect method offers several advantages:

- it spares the necessity to fit the likelihoods,

- in the multi-dimensional case, contrary to the naïve Bayes classifier, the independence assumption is not necessary, because the logistic regression automatically takes care of the correlation between features,

- by construction, the log LR is defined by a hyperplane, and thus robust with respect to the equal variance assumption, and to outliers or sparse data far from the boundary,

- in the forensic context, since the log LR is directly proportional to a$^T$ d + b, the logistic allows a convenient and interpretable calibration of the dissimilarity score d, and a fusion of scores in the multidimensional case [18].

On the other hand, the indirect method might suffer from:

- a reduced flexibility since it amounts to assuming single-Gaussian distributions,

- a possibly important computation time in the case of high-dimensional evidence and of a distance of the same dimension.

These advantages and disadvantages will be examined and discussed on the example of hand odor evidence using a large panel of subjects.

## 2.4. Dataset description

A panel of 534 volunteers was set up which gathers 218 men and 316 women aged 7 to 94 years (median 28, interquartile interval [22 ; 48]), see Table 1 for the detailed composition in terms of gender and age. Note that this composition does not aim at reflecting that of a precise target population, such as one which is more likely to commit a crime, but to be as representative of the diversity of odors as possible. As a matter of fact, criminal investigations also often necessitate to look for victims, or to discriminate between traces from different people present at a crime scene, including those of victims or witnesses, who might be women as well as men, children or seniors as well as middle-aged adults. All data were completely anonymized prior to analysis, and no personal information was stored.

The goal was here to identify the subjects by their hand odor, whose volatile profile was shown to display a between-subject variability which is sufficient for differentiation [21, 5]. Also, in the forensic context, the hands have the advantage to be more likely to be directly in contact with objects at a crime scene, and to be easier to sample during a police interrogation.

The volatile profiles were obtained by a direct sampling procedure using identical sample collection kits of 4 small polymer strands that the subjects were asked to rub together in their hands for 15 minutes. The polymer strands were thermodesorbed, and the concentrated substances were separated by comprehensive bidimensional gas chromatography (GCxGC) coupled with mass spectrometry (MS). The sampling method and the optimization of the GCxGC-MS analysis were extensively described in [22-23]. Data were acquired, converted to .mzXML files with GC Real Time Analysis 4.20 (Shimadzu software), and then processed with MatlabTM (Natick, MA, USA) version 9.6.0.1150989 (R2019a), its Statistics and Machine Learning Toolbox version 11.5 and its Bioinformatics Toolbox version 4.12.

Using a "home-made" Matlab script [24], the preliminary manual processing of 25 chromatograms obtained on 3 subjects between 23 and 26 years old of both genders sampled several times at different time instants enabled us to draw up a first list of several hundreds of peaks. A library was built to store their retention times, their linear retention index, their mass spectrum, and the name of the corresponding compound when it could be identified using the NIST library. Indeed, if the availability of its mass spectrum is compulsory, a compound does not need to be formally identified for the comparison of chromatograms. We also checked whether compounds described in the literature as constituents of the human hand odor were present in this library, otherwise they were included. The library was then continuously enriched as the panel was increased with compounds potentially relevant to human hand odor because of their empirical frequency in new samples. This work led us to a customized library of 741 compounds, which were looked for in each chromatogram. As a result, each sample was characterized by the peak areas of 741 compounds.

In order to compensate for uncontrolled variations of the total area of the chromatograms, the sum of these areas was normalized in logarithmic scale to unit value, see Table 2 for a comparison of the reproducibility of the data without normalization, and normalization in scalar and logarithmic scales. Not knowing whether all 741 compounds are really relevant for identification, such a normalization across all compounds might be questionable. Thus, the possibility to avoid the problem by working on the dichotomized areas, i.e. 1 if the compound is present, or 0 if it is absent from the sample, was also investigated. Also, this approach might be of interest for forensic identification problems dealing with intrinsically binary features, such as gradient, structural and concavity (GSC) binary features in handwriting identification [25]. In the following, we refer to these traces of 741 features, continuous or dichotomized, as "odor traces".

As stated above, the subjects were sampled in quadruplicate, but due to unavoidable mishaps with some samples (like accidently dropping a polymer on the floor during sampling) and to chromatographic problems (such as failures of the cryogenic modulator), 1690 odor traces were obtained for the 534 subjects (44 were sampled once, 77 twice, 160 three times, and the remaining 253 subjects four times, leading to an average of 3.2 odor traces per subject). This data set was split into a calibration set for training and validation, and an independent test set for performance estimation. Since gender [26] and age [27] are known to impact odor traces, the split was made so as to respect the gender proportions, with subject of all ages in the two sets, and odor traces of the same subject being put in the same set. As a result, the calibration set comprises 412 subjects and their 1 299 odor traces (corresponding to 1 594 $H_{ss}$ and 841 457 $H_{ds}$ pairs), and the test set comprises the remaining 122 subjects and their 391 odor traces (leading to 481 $H_{ss}$ and 75 764 $H_{ds}$ pairs). The way the odor traces distribute between calibration and test set can be grasped through the Principal Component Analysis (PCA) of Figure 1.

*2.5. Implementation*

Three methods are implemented:

1) the direct method using a scalar distance between odor traces,

2) the indirect method using a scalar distance between odor traces,

3) the indirect method using a vectorial distance, i.e. a distance on each odor compound.

Note that, given the large dimension of the problem (n=741) and the known correlations between features, we did not attempt to implement the direct method using a vectorial distance (i.e. a scalar distance on each feature and the naïve Bayes classifier).

2.5.1. Distances between two odor traces

Let $x_i$ denote the n-vector representing odor trace i. A standard choice of scalar dissimilarity measure between odor traces i and j is the Euclidian distance between vectors $x_i$ and $x_j$, i.e.:

$$d_{Euclid}(x_i, x_j) = \sqrt{\sum_{k=1}^{n}(x_i^k - x_j^k)^2} \qquad (7)$$

But this distance is not robust with respect to shifts and linear transformations of the features. Thus, a more appropriate distance would be Pearson's linear correlation based distance:

$$d_{Pearson}(x_i, x_j) = 1 - \frac{\sum_{k=1}^{n}(x_i^k - \overline{x}_i)(x_j^k - \overline{x}_j)}{\sqrt{\sum_{k=1}^{n}(x_i^k - \overline{x}_i)^2 \sum_{k=1}^{n}(x_j^k - \overline{x}_j)^2}} \qquad (8)$$

However, Pearson's correlation is sensitive to non-linearities, whereas Spearman's non-parametric correlation coefficient on the ranks is able to capture monotonic nonlinear associations as well as linear ones [28]. Thus, the Spearman correlation based distance (same as Equation (8) with the $x_i^k$ and $x_j^k$ replaced by their ranks) is expected to be more robust with respect to nonlinear variations of the peak areas. The three distances were compared in a previous study where Spearman's correlation based distance indeed clearly outperformed the two other distances [29]. Therefore, the direct and indirect method using a scalar distance have been implemented with Spearman's correlation based distance.

The chosen vectorial distance for the indirect method using a vectorial distance is simply the vector of the absolute differences between feature values:

$$d_{vectorial}(x_i, x_j) = \left[ |x_i^1 - x_j^1| \ldots |x_i^n - x_j^n| \right]^T \qquad (9)$$

### 2.5.2. Estimation of the likelihoods for the direct method

The calibration set was used to build pairs of odor traces of same and different sources, and to compute their distances. The empirical densities being essentially uni- or bimodal (see Figure 2), they were best fitted with a two-Gaussian mixture distribution, using Matlab's function "fitgmdist", leading to estimates of the likelihoods $f(d|H_{ss})$ and $f(d|H_{ds})$.

When performing feature selection (see section 2.5.5 below), the same-source and different-source distributions vary, depending on the subset of features. However, checking the optimality of the two-Gaussian mixture for all subsets of features during the feature selection process would be too time consuming. Nevertheless, the convergence of the fits was checked for, each fit being repeated three times with a different random set of initial parameters in order to retain the fit with the largest likelihood, and the optimality of the fit obtained with the selected number of features was tested. Of course, as stated in section 2.3, the necessity to fit the likelihoods and to optimize these fits, or to suffer suboptimal fits, is the first disadvantage of the direct method.

### 2.5.3. Estimation of the logistic model for the indirect methods

The logistic regression model of Equation (3) with parameters $\theta = [a^T \ b]^T$ was fitted through maximum likelihood, by minimizing the cross-entropy cost function with Matlab's function "glmfit".

### 2.5.4. Likelihood ratio and performance estimation

For the direct method, the LR was evaluated using the estimates of the likelihoods $f(d|H_{ss})$ and $f(d|H_{ds})$ and Equation (2), as a function of the distance d. For the indirect methods, the LR was obtained from the fitted logistic regression and Equation (6), and plotted as a function of the distance d or of the score $a^T d + b$, depending on d being scalar of vectorial.

Since there is no true reference for the LR, the performance of the different methods was evaluated by estimating the posterior probability $P(H_{ss}|d)$ according to Equations (1) and (5) for the direct and indirect methods respectively, and by performing a binary classification using equal prior probabilities ($P(H_{ss}) = P(H_{ds}) = 0.5$). Varying the decision threshold on $P(H_{ss}|d)$, the sensitivity and the specificity were estimated on the calibration and test sets, and used to compute the corresponding areas under the receiver operating characteristic "ROC" curve (AUC) [30]. The performance was further characterized by the sensitivity Sn and specificity Sp maximizing Youden's index [31], i.e. Sn + Sp −1.

2.5.5. Feature selection

In a previous study [29], improved results were obtained using feature selection. The idea is to retain the features that contribute the most to the difference between distance densities under $H_{ss}$ and $H_{ds}$. Given the large number of features (odor compounds) and the large size of the data set, a filter approach to this selection was chosen. Filter approaches are based on a statistical measure of the difference between the two densities for each feature. They are hence independent from the main algorithm (direct or indirect method), as opposed to the so-called wrapper approaches, which sequentially evaluate the relevance of each feature subset based on the performance of the whole procedure, i.e. feature selection together with main algorithm [32]. Since filter approaches consider the features independently, they might retain redundant features, but in turn this gives them more robustness and, most importantly, they require less computation time. For each feature, considering the absolute values of the differences for the $H_{ss}$ and $H_{ds}$ pairs, we chose Wilcoxon's non-parametric test statistic as statistical measure in the case of continuous features, and Fisher's exact test statistic in the case of dichotomized features. Then, the features were ranked in decreasing order of the one-sided p-value of the test, which is a one-sided test since smaller distances between features under $H_{ss}$ than under $H_{ds}$ are sought for. The number of features maximizing the AUC was estimated on the calibration set using 3-fold cross-validation. The cross-validation partitions were randomly chosen with the constraint that the odor traces of the same subject were put in the same partition. Note that cross-validation also enabled us to estimate the uncertainty on the AUCs through the mean standard deviation on the three partitions.

**3. Results and discussion**

The three methods are first evaluated using all the features of the odor traces (baseline comparison) and then, the possibility to further improve their performance using feature selection is investigated.

*3.1. Baseline comparison of the three methods (without feature selection)*

The results obtained with the three methods on the calibration and test sets are summarized in Tables 3 and 4 for dichotomized and continuous features respectively. As a first remark, the performance of the direct and indirect methods using a scalar distance in terms of AUC and of sensitivity and specificity are almost identical, for both dichotomized and continuous features. Thus, the higher flexibility of the direct method does not increase the performance. On the contrary, its lack of robustness can be visualized on Figure 2 depicting the posterior probability and the LR obtained with the dichotomized features: due to the larger variance of the likelihood under $H_{ss}$, the posterior probability and the LR, instead of being monotonous, start to increase with the distance at some point (d ≈ 0.7). Whereas whatever the situation with the indirect method, posterior probability and LR always decrease with d, see Figure 3 depicting the posterior probability and the LR obtained with indirect method, this time on the continuous features.

Also noteworthy, the performance obtained with the indirect method using a vectorial distance is significantly better than those of the methods working with a scalar distance: the AUC on the calibration and test sets jumps from 91-92% to 97-98%, the standard deviation of the AUC being estimated at 0.7% using 3-fold cross-validation on the calibration set. The distributions of the score ($a^T d + b$) resulting from the logistic regression, the regression itself, the posterior probability and the LR are shown in Figure 4. The only drawback lies in the increased, but perfectly tractable computational cost (10 minutes instead of a few seconds, on a 4,2 GHz Intel Core i7).

Finally, the binarization of the features decreases the performance, but only marginally (the AUC is decreased by ≈ 1%). Note that, in this precise case where the features quantify the amount of odor compounds, this could be due to the fact that the normalization of the compound proportion uses all these compounds whereas it is not known whether they are all relevant. Note also that the normalization was improved by performing it in the logarithmic scale rather than in the linear scale (the former improving the reproducibility, see Table 2), with which continuous features did not outperform dichotomized features, as shown in a previous study [29]. Finally, other normalization methods specific to GCxCG-MS data might advantageously be investigated [Chen et al. 2017], but are outside the scope of this paper.

*3.2. Comparison of three methods with feature selection*

The number of selected features using the filter approach is reported in Tables 5 and 6, together with the corresponding results on the calibration and test sets, for dichotomized and continuous feature respectively.

Again, there is almost no difference in performance between the direct and indirect methods with a scalar feature, be it on dichotomized or continuous features. In terms of AUC, the selection is more efficient on binary features than on continuous ones (94.4% with selection instead of 91.5% without for dichotomized features, 93.5% instead of 93.0% for continuous features, on the test set), with an important reduction of the number of the dichotomized features (267 instead of 741), and a moderate one for continuous features (535 instead of 741). Note also that in both cases, this increased performance benefits the specificity, which is highly desirable in a forensic application (it is crucial in this context not to reject the different-source hypothesis, i.e. the defense hypothesis, when in fact it is true).

For both dichotomized and continuous features, the indirect method using a vectorial distance is again significantly better than the previous ones (AUCs around 97-98% instead of 94-95% on both calibration and test sets), with a similar number of selected features (440 for dichotomized features, 500 for continuous ones). However, in both cases, the parsimony due to feature selection does not increase the performance as compared to the baseline method, it is quasi-identical with and without selection. In return, this testifies to a robustness of the indirect method with respect to possibly irrelevant features. And of course, not to have to perform the selection spares computation time.

### 3.3. Discussion of the choice of equal priors

In this manuscript, the methods are compared in terms of AUC, sensitivity and specificity. In the case of the indirect methods, the regression being obtained by fitting a logistic function to the data, whatever the prior probabilities $P(H_{ss})$ and $P(H_{ds})$, the posterior probability $P(H_{ss}|d)$ given by Equation (5) is also a logistic function. Thus, when the threshold on $P(H_{ss}|d)$ is varied from 1 to 0, the same ROC curve is described, whose AUC only depends on the distance distributions under $H_{ss}$ and $H_{ds}$: only the threshold maximizing Youden's index changes.

With the direct method, the choice of the prior has an influence on the shape of the posterior probability, so that the threshold on $P(H_{ss}|d)$ can possibly be varied in a different interval (see Figure 3 where $P(H_{ss}|d)$ never reaches 0 for example). However, in practice, there is no influence on AUC, sensitivity and specificity because, again, the AUC depends essentially on the distance distributions under $H_{ss}$ and $H_{ds}$. The only palpable change is on the threshold yielding the best compromise between sensitivity and specificity, threshold which adjusts to $P(H_{ss})$ by roughly following it.

Thus, the assumption of equal prior probabilities has practically no impact on the LR estimate.

*3.4. Limitations*

From a practical point of view, our work suffers several limitations for a real-world forensic application. First, for practical reasons, the subjects were sampled at a single time point, so that the variability of the data is essentially due to the analytical variability. Second, the chromatograms being compared are of the same nature, i.e. obtained on samples provided by directly sampling the subjects (with contact with the subjects' hands) whereas in real life, the unknown source sample will be obtained indirectly from an object on the crime scene (without contact with the subject). Third, the odor collected on the crime scene might be contaminated by other odors, from the environment or from other people. A study focused on mixtures of odors, contaminations, and weathered traces has not been carried out yet but is considered. However, despite these controlled conditions, the PCA of Figure 1 and the statistics of Table 2 show that the data is already of limited reproducibility, so that the good results we have obtained are encouraging concerning the robustness of the best method to more realistic sampling conditions.

From a methodological point of view, the proposed methods are based on a common source scenario, where it is asked whether the two traces originate from the same source or from different sources without specifying which sources are considered, and not on a specific source scenario, where the question is whether the two traces stem specifically from the known source [34]. The problem of the common source scenario is that it does not take account of the typicality of the source, contrary to recommendations for a better estimation of the strength of evidence through the LR [18, 7, 35]. But to implement a specific source scenario, a number of traces from the known source are needed in order to be able to estimate the distribution under $H_{ss}$ (for the direct method) or to discriminate between the $H_{ss}$ and $H_{ds}$ populations (for the indirect methods), which is quite unpractical when dealing with human hand odor, and not feasible at this stage of the study (at most four usable odor traces were obtained, for only 253 subjects among the 534).

**4. Conclusion**

To summarize, the advantages expected from an indirect method are fully obtained, in particular the dispensation to parameterize the likelihoods, and the robustness with respect to differences in their variance and/or to possible outliers. Moreover, an increase in performance of the indirect method as compared with the direct one is not obtained with a scalar distance between odor traces, but when using the vector of the distances between each feature of the odor traces. This improvement was not really expected, because, especially in the forensic context, it is often advocated to convert multivariate data to a univariate datum summarizing the relationship between features. Finally, the indirect method with a vectorial distance proves also robust with respect to potentially irrelevant features since removing them does not modify the performance, an appealing quality for dealing with traces which are not yet solidly characterized, such as odor traces.

**Table 1.** Panel composition in terms of gender and age.

| age / gender | 7-17 | 18-64 | 65-94 | total |
|---|---|---|---|---|
| man | 18 | 182 | 18 | 218 |
| woman | 28 | 268 | 20 | 316 |
| total | 46 | 450 | 38 | 534 |

**Table 2.** Repeatability of the odor trace features (peak areas), without and with two normalizations, estimated on the 490 subjects sampled at least twice (IQI stands for interquartile interval).

| Normalization | Median relative standard deviation [IQI] in % |
|---|---|
| None | 61.0 [32.1 ; 78.0] |
| In scalar scale | 56.1 [31.9 ; 74.2] |
| In logarithmic scale | 33.2 [17.9 ; 48.8] |

**Table 3.** Baseline comparison between the three methods on the calibration and test sets, using dichotomized features, in terms of AUC, sensitivity (Sn) and specificity (Sp) for the threshold maximizing Youden's index, all in %.

|  | Calibration | | | | Test | | | |
|---|---|---|---|---|---|---|---|---|
| Method | AUC | threshold | Sn | Sp | AUC | threshold | Sn | Sp |
| Direct | 91.2 | 0.43 | 80.6 | 91.3 | 91.4 | 0.54 | 78.4 | 94.9 |
| Indirect scal. d | 91.2 | 0.62 | 80.6 | 91.3 | 91.5 | 0.75 | 78.4 | 94.9 |
| Indirect vect. d | 98.5 | 0.54 | 93.4 | 96.3 | 97.1 | 0.56 | 91.1 | 94.2 |

**Table 4**. Baseline comparison between the three methods on the calibration and test sets, using continuous features, in terms of AUC, sensitivity (Sn) and specificity (Sp) for the threshold maximizing Youden's index, all in %.

|  | Calibration | | | | Test | | | |
| --- | --- | --- | --- | --- | --- | --- | --- | --- |
| Method | AUC | threshold | Sn | Sp | AUC | threshold | Sn | Sp |
| Direct | 92.1 | 0.44 | 81.1 | 92.5 | 93.0 | 0.50 | 81.3 | 94.6 |
| Indirect scal. d | 92.1 | 0.64 | 81.1 | 92.5 | 93.0 | 0.73 | 81.3 | 94.6 |
| Indirect vect. d | 98.9 | 0.58 | 94.4 | 97.0 | 97.8 | 0.57 | 91.3 | 95.1 |

**Table 5**. Comparison between the three methods on the calibration and test sets, using dichotomized features, in terms of AUC, sensitivity (Sn) and specificity (Sp) maximizing Youden's index, with selection of the number of features among the 741 by cross-validation of the calibration set (AUC-CV3 the mean 3-fold cross-validation AUC on the calibration set, and #feat. denotes the number of features selected by cross-validation).

|  |  |  | Calibration | | | Test | | |
| --- | --- | --- | --- | --- | --- | --- | --- | --- |
| Method | AUC-CV3 | #feat. | AUC | Sn | Sp | AUC | Sn | Sp |
| Direct | 94.4 | 267 | 94.5 | 80.9 | 94.7 | 94.4 | 84.6 | 92.8 |
| Indirect scal. d | 94.4 | 267 | 94.5 | 80.9 | 94.7 | 94.4 | 84.6 | 92.8 |
| Indirect vect. d | 96.4 | 440 | 97.6 | 89.5 | 97.5 | 97.0 | 91.3 | 94.3 |

**Table 6.** Comparison between the three methods on the calibration and test sets, using continuous features, in terms of AUC, sensitivity (Sn) and specificity (Sp) maximizing Youden's index, with selection of the number of features among the 741 by cross-validation of the calibration set (AUC-CV3 the mean 3-fold cross-validation AUC on the calibration set, and #feat. denotes the number of features selected by cross-validation).

|                  |         |        | Calibration |      |      | Test |      |      |
|------------------|---------|--------|------|------|------|------|------|------|
| Method           | AUC-CV3 | #feat. | AUC  | Sn   | Sp   | AUC  | Sn   | Sp   |
| Direct           | 93.1    | 535    | 93.1 | 81.3 | 93.3 | 93.5 | 82.3 | 95.5 |
| Indirect scal. d | 93.1    | 535    | 93.1 | 81.3 | 93.3 | 93.5 | 82.3 | 95.5 |
| Indirect vect. d | 97.0    | 500    | 98.3 | 91.6 | 97.3 | 97.7 | 91.7 | 94.9 |

**FIGURE CAPTIONS**

**Figure 1.** PCA of the data set showing the distribution of the odor traces between calibration and test sets. The PCA was performed on the covariance matrix of the continuous features, i.e. normalized in logarithmic scale.

**Figure 2**. Results of the direct method with the scalar distance d on dichotomized feature, as functions of d, on the calibration set. a) Empirical distribution of d under $H_{ss}$ (1 594 pairs), and estimated density (mixture of 2 Gaussians). b) Empirical distribution of d under $H_{ds}$ (841 457pairs), and estimated density (mixture of 2 Gaussians). c) Posterior probability of $H_{ss}$ obtained using Bayes' formula with equal priors. d) Likelihood ratio.

**Figure 3**. Results of the indirect method with the scalar distance d on continuous features, as functions of d, on the calibration set. a) Empirical distribution of d under $H_{ss}$ (1 594 pairs). b) Empirical distribution of d under $H_{ds}$ (841 457pairs). c) Logistic regression (dotted line), and deduced posterior probability of $H_{ss}$ with equal priors (continuous line). d) "Likelihood ratio" $\exp(a^T d + b)$ corresponding to the logistic regression (dotted line), and likelihood ratio (continuous line).

**Figure 4**. Results of the indirect method with the vectorial distance d on continuous features, as functions of the score $a^T d + b$, on the calibration set. a) Empirical distribution of the score $a^T d + b$ under $H_{ss}$ (1 594 pairs). b) Empirical distribution of the score $a^T d + b$ under $H_{ds}$ (841 457pairs). c) Logistic regression (dotted line), and deduced posterior probability of $H_{ss}$ with equal priors (continuous line). d) "Likelihood ratio" $\exp(a^T d + b)$ corresponding to the logistic regression (dotted line), and likelihood ratio (continuous line).

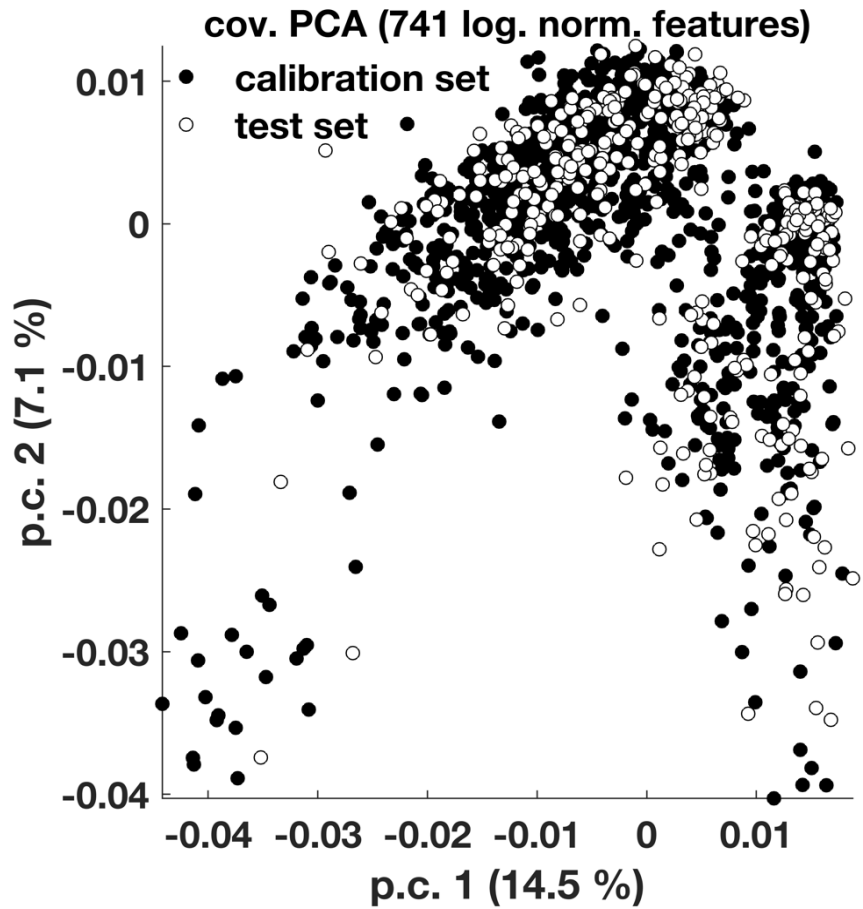

**Figure 1.** PCA of the data set showing the distribution of the odor traces between calibration and test sets. The PCA was performed on the covariance matrix of the continuous features, i.e. normalized in logarithmic scale.

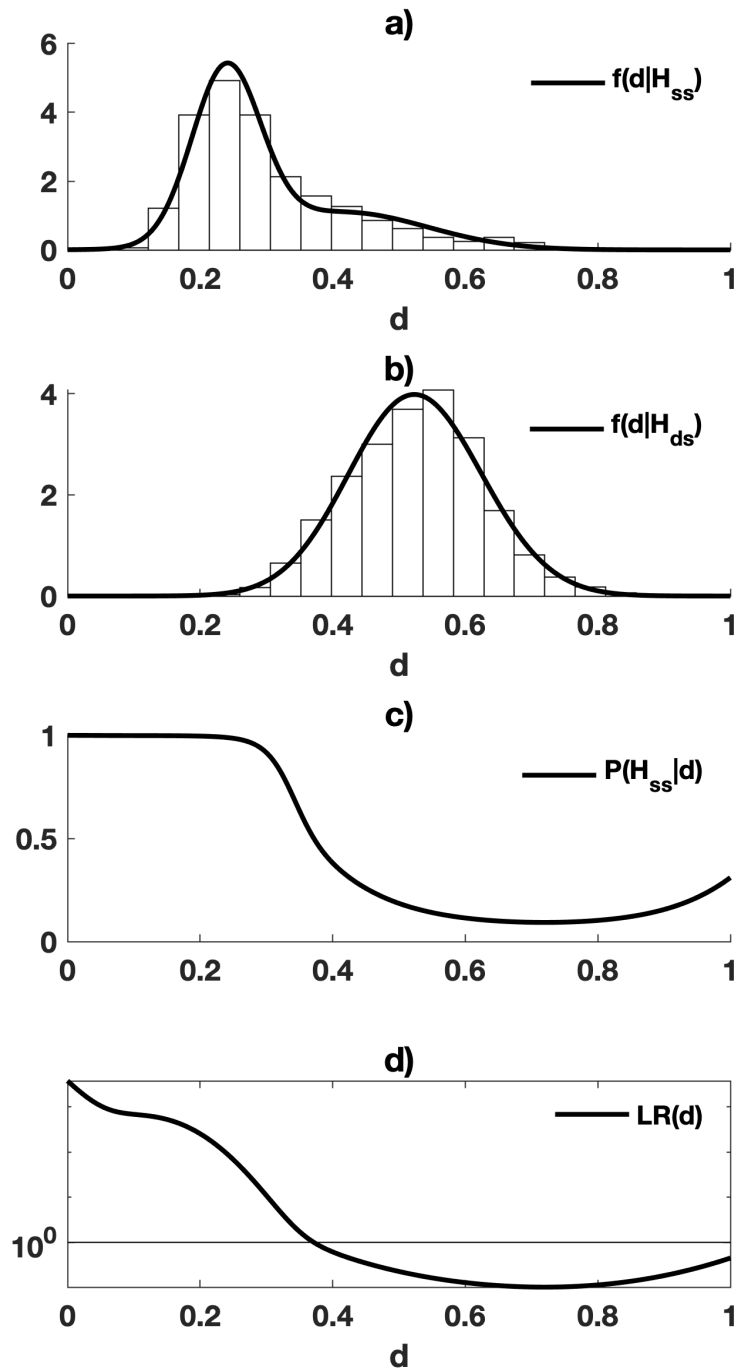

**Figure 2.** Results of the direct method with the scalar distance d on dichotomized feature, as functions of d, on the calibration set. a) Empirical distribution of d under $H_{ss}$ (1 594 pairs), and estimated density (mixture of 2 Gaussians). b) Empirical distribution of d under $H_{ds}$ (841 457pairs), and estimated density (mixture of 2 Gaussians). c) Posterior probability of $H_{ss}$ obtained using Bayes' formula with equal priors. d) Likelihood ratio.

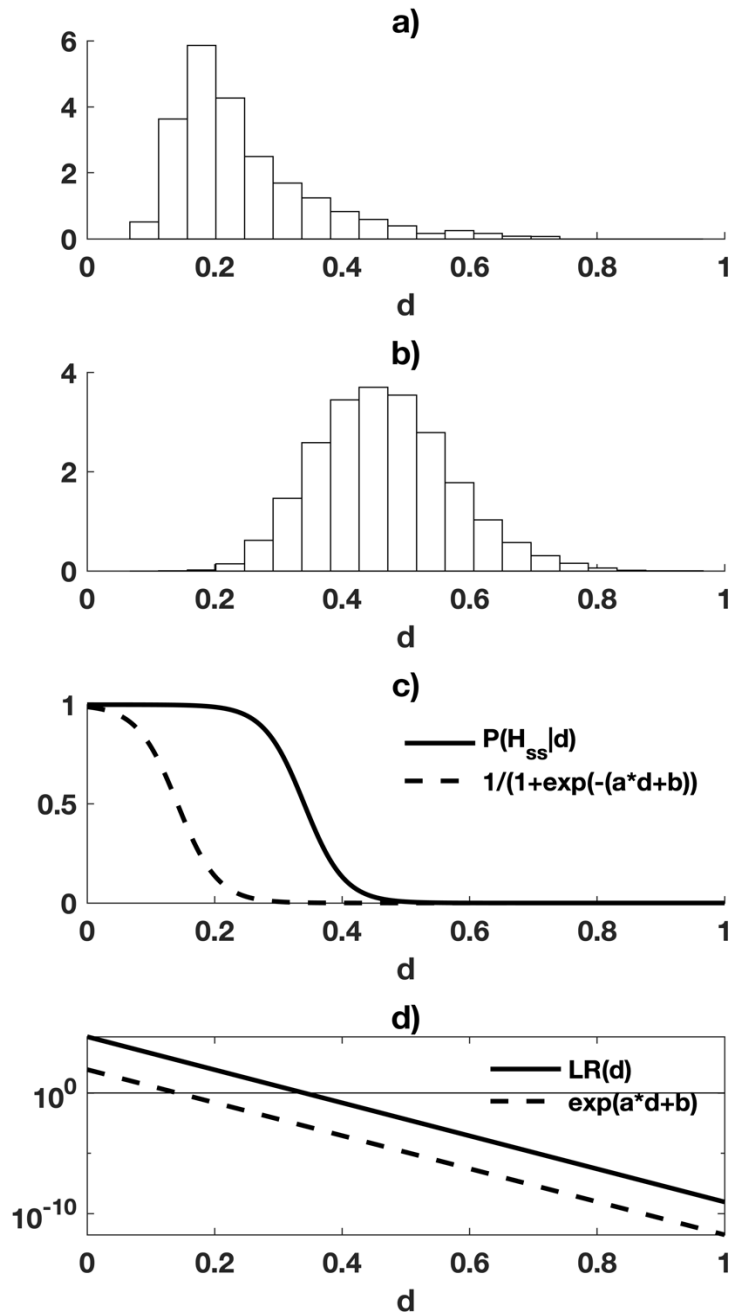

**Figure 3**. Results of the indirect method with the scalar distance d on continuous features, as functions of d, on the calibration set. a) Empirical distribution of d under $H_{ss}$ (1 594 pairs). b) Empirical distribution of d under $H_{ds}$ (841 457pairs). c) Logistic regression (dotted line), and deduced posterior probability of $H_{ss}$ with equal priors (continuous line). d) "Likelihood ratio" $\exp(a^T d + b)$ corresponding to the logistic regression (dotted line), and likelihood ratio (continuous line).

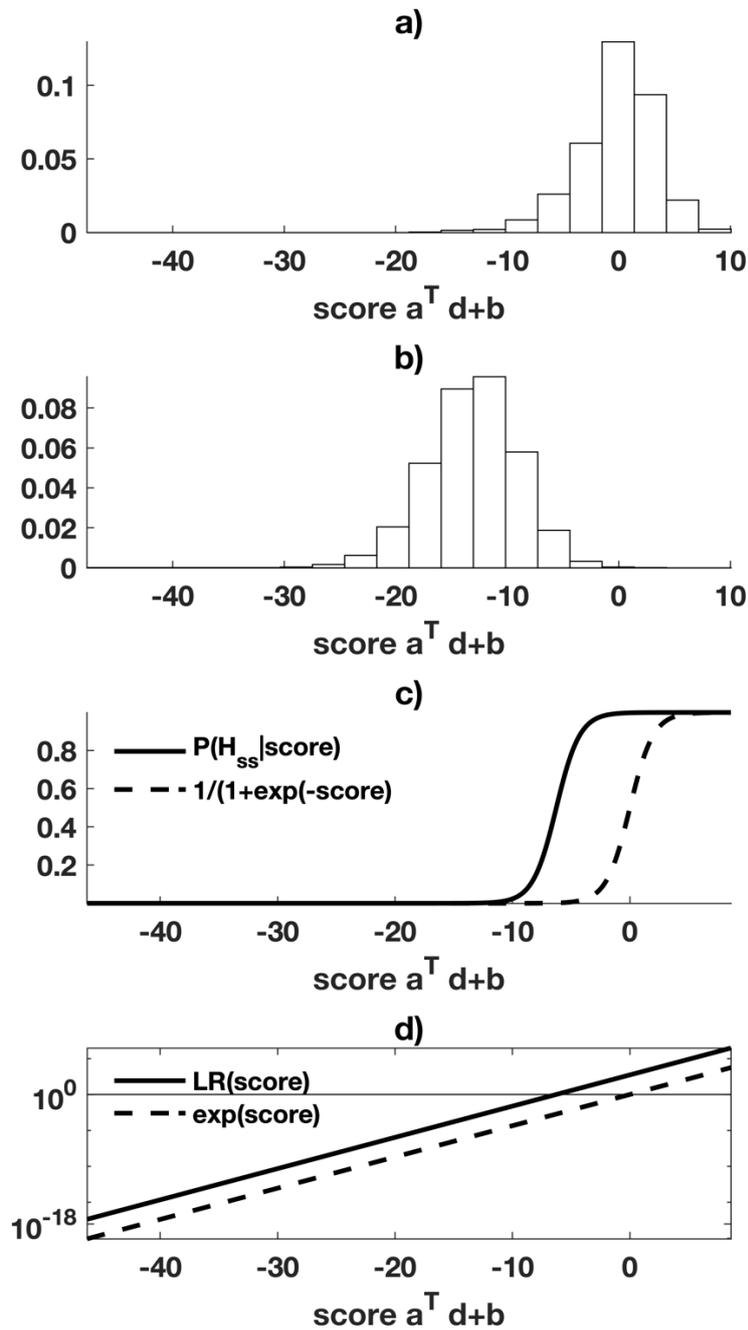

**Figure 4**. Results of the indirect method with the vectorial distance d on continuous features, as functions of the score $a^T d + b$, on the calibration set. a) Empirical distribution of the score $a^T d + b$ under $H_{ss}$ (1 594 pairs). b) Empirical distribution of the score $a^T d + b$ under $H_{ds}$ (841 457 pairs). c) Logistic regression (dotted line), and deduced posterior probability of $H_{ss}$ with equal priors (continuous line). d) "Likelihood ratio" $\exp(a^T d + b)$ corresponding to the logistic regression (dotted line), and likelihood ratio (continuous line).